\documentclass[]{interact}
\usepackage{epstopdf}% To incorporate .eps illustrations using PDFLaTeX, etc.
\usepackage{subfigure}% Support for small, `sub' figures and tables
\graphicspath{ {./Images/} }
\usepackage{natbib}% Citation support using natbib.sty
\bibpunct[, ]{(}{)}{;}{a}{}{,}% Citation support using natbib.sty
\theoremstyle{plain}% Theorem-like structures provided by amsthm.sty

\theoremstyle{definition}

\theoremstyle{remark}

\usepackage{utfsym}
\usepackage{multirow}

\begin{document}
\articletype{RESEARCH ARTICLE}
\title{An LSTM-Based Predictive Monitoring Method for Data with Time-varying Variability}
\author{
	\name {J. QIU \textsuperscript{a,c}\thanks{CONTACT J. QIU. Email: j.qiu@uva.nl}  and Y. Lin \textsuperscript{a} and I. M. Zwetsloot \textsuperscript{a,b,c}}
	\affil {\textsuperscript{a} Department of Systems Engineering, City University of Hong Kong;
		\textsuperscript{b} School of Data Science, City University of Hong Kong;
		\textsuperscript{c} Department of Business Analytics, Amsterdam Business School, University of Amsterdam}
}

\maketitle

\begin{abstract}
The recurrent neural network and its variants have shown great success in processing sequences in recent years. However, this deep neural network has not aroused much attention in anomaly detection through predictively process monitoring. Furthermore, the traditional statistic models work on assumptions and hypothesis tests, while neural network (NN) models do not need that many assumptions. This flexibility enables NN models to work efficiently on data with time-varying variability, a common inherent aspect of data in practice. 
This paper explores the ability of the recurrent neural network structure to monitor processes and proposes a control chart based on long short-term memory (LSTM) prediction intervals for data with time-varying variability. The simulation studies provide empirical evidence that the proposed model outperforms other NN-based predictive monitoring methods for mean shift detection. The proposed method is also applied to time series sensor data, which confirms that the proposed method is an effective technique for detecting abnormalities.
\end{abstract}

	\section{Introduction}
		The rapid development in information technology and the increasing use of sensors and computers make it possible to measure the performance of (industrial) systems in real-time. This data is becoming an integral part of health management and decision-making. One aspect of health management is using sensor data for continuous and real-time monitoring of systems in order to detect changes (for example, indicating possible system failures). One promising method for anomaly detection is the application of neural network (NN) models, which are well-known for their ability to effectively deal with massive sensor data. 
		
		Hence, the development of anomaly detection and predictive monitoring methods based on NN models has attracted attention in the literature. For example, \citet{pugh1989synthetic} proposed a NN based on linear separation of the data problem, and studied their method via simulation experiments and on three labelled datasets. Another approach is to use pattern recognition. For example, \cite{de2018identification} proposed three control chart pattern recognition systems for three special stationary time series processes. \citet{yu2019selective} proposed a stacked de-noising auto-encoder (SDAE) model-based approach for feature learning and high dimensional process pattern recognition. \citet{fuqua2020cost} developed a cost-sensitive convolution neural network (CSCNN) for imbalanced process pattern recognition. And \citet{maged2022recognition} proposed a CNN for variable input sizes (VIS-CNN) to monitor signals with different lengths. \citet{arkat2007artificial} proposed an ANN-based CUSUM chart to monitor multivariate AR(1) processes. \citet{chen2019deep} proposed an RNN-based residual charts method to detect mean shifts in AR processes. The reader is referred to \citet{tran2022application} for a more comprehensive list of papers on this general topic. An alternative to NN models is conventional time series models. However, it is well documented that Artificial Neural Network (ANN) performs better than ARIMA in forecasting stock price \citep{wijaya2010stock,adebiyi2014comparison}, and \citet{chen2011comparison} obtained the same results in short-term wind speed forecasting. 
		
		Nevertheless, there still exist some challenges in applying NN for change detection in real-time. The existing methods require 1) numerous and reliable labelled sensor data in training and 2) that the training data has stable variability over time. 
		
		Reliable labels for sensor data in real life are not always straightforward to collect. They are often very expensive to obtain, as it has to be collected manually. In addition, in our experience, the engineers might not precisely know the current operating condition of the system hence making labelling impossible. To deal with the challenge of unlabelled data, current work usually obtains labels from simulations of statistic models \citep{de2018identification,yu2019selective,fuqua2020cost,maged2022recognition}.
		However, this leads to another challenge: in-control behaviours may be too complex to simulate accurately since the sensor data are often noisy and non-stable. This means that the training sample establishment is somewhat troublesome, as \citet{weese2016statistical} mentioned. In addition, in practice, the out-of-control patterns may be more complicated than the simulated patterns, or they may even unknown patterns.
		
		The second challenge: obtaining a stable training data set with constant variability over time. In certain applications, this is impossible, as time-varying variability is often a inherent aspect of the process. In particular, seasonality can influence the variability of the process under study, like the variability of daily traffic flows changes periodically. The variability in vibration may be affected by workload of the component. For example, \citet{wang2021changepoint} consider sensor data that shows time-varying variability. When data or data collection processes are affected by the environment and these dynamic external influences are difficult to control the data may show time-varying variability. Existing monitoring methods, assume stable variability over time and thus ignore this heteroscedasticity, which in turn may heavily affect the detection performance. 
		
		In this work, we introduce a novel long short-term memory (LSTM)-Based Anomaly Detection framework for the scenario where the data has time-varying noise. We choose to use an LSTM model to deal with the first challenge as LSTM models are trained by taking the value of the next time stamp as the label. To effectively deal with data with time-varying noise we explicitly model the data uncertainty and predict it using an ANN. 

		Therefore, the main contributions of our work is the development of real-time anomaly detection framework, that:
			\begin{itemize}
				\item Does not rely on labelled data as the LSTM model is trained on the data itself.
				\item Allows for time-varying variability in the training data through a data uncertainty quantification module.
			\end{itemize}
		
		We study the performance of our method and compare it with an existing method through simulation and an ablation study. In addition, we implement our method on two case studies.
		The first case focuses on vibration data, and shows that our proposed monitoring method is effective in detecting changes in vibration data which show time-varying variability and for which there are no labels available. The second case study shows that our method is very flexible as this data has a very different structure to the first case study.
	
		The remaining parts of this paper are organized as follows. The next section gives an introduction to the technique used in this paper. Our proposed model is introduced and discussed in Section 3. The theoretical performance of our model and simulation experiments, including an ablation study, are presented in Section 4. We present two implementations of our model on escalation vibration and energy data in Section 5. In the final section, the conclusion of this paper is drawn and future work is discussed.

	\section{LSTM Model}
		This section introduces a brief overview of the LSTM models utilized in our approach.
		The LSTM model is a recurrent neural network and a variant of the Elman RNN, the most well known RNN \citep{elman1990finding}. RNN structure models have attracted great attention in time series prediction in sequential learning like text generating \citep{sutskever2011generating}, speech recognition \citep{graves2013speech} and so on. However, learning long-term dependencies is challenging for the Elman RNN. The LSTM models proposed by \citet{hochreiter1997long} are not affected by this deficiency by introducing a memory cell in the hidden layer. 
		The most famous and commonly used LSTM model was proposed by \citet{gers2000learning} which adds a forget gate in the LSTM unit to reset the LSTM when the contents in the memory cells become irrelevant. 
		
		Thus, in this work, an LSTM unit is a recurrent architecture with an input, output, and forget gate, and the LSTM model consists of a sequence of LSTM units and a fully connected layer.
		The following formulas describe its computation.
			\begin{equation}\label{equa:1}
				i_{t} = \sigma(x_{t}U^i+h_{t-1}W^i)
			\end{equation}
			\begin{equation}\label{equa:2}
				o_t = \sigma(x_{t}U^{o}+h_{t-1}W^{o}) 
			\end{equation}
			\begin{equation}\label{equa:3}
				f_{t} = \sigma(x_{t}U^{f}+h_{t-1}W^{f}) 
			\end{equation}
			\begin{equation}\label{equa:4}
				\tilde{C}_{t}=\mathrm{tanh}(x_{t}U^{g}+h_{t-1}W^{g})
			\end{equation}
			\begin{equation}\label{equa:5}
				C_{t} = \sigma(f_{t}*C_{t-1}+i_{t}*\tilde{C}_{t}) 
			\end{equation}
			\begin{equation}\label{equa:6}
				h_{t} = \mathrm{tanh}(C_{t}*o_{t}) 
			\end{equation}	
		where \(i_{t}\), \(o_{t}\), and \(f_{t}\) denote input, output, and forget gate, respectively.
		\(U\) and \(W\) denote the weights for the input and hidden layers.	Sigmoid \(\sigma\) and hyperbolic tangent \(\mathrm{tanh}\) are the non-linear activation functions. \(\tilde{C}_{t}\) represents the candidate value of the memory block, and \(C_{t}\) is the current value of the memory block. The output of an LSTM unit is \(h_{t}\), as shown in (\ref{equa:6}). And a sequence of outputs from LSTM units will be the input to a fully connected layer. After that, the result \(y_{t} = \vec{w} \cdot \vec{h}+b\) is the final output of the model.
	\section{Proposed Model}
	In this section, we introduce our proposed model. First Section 3.1  provides the exposition of the assumed data structure. Details of our proposed model are discussed in Sections 3.2 and 3.3. 
	
		\subsection{Data}
		Denote the observed time series by \(\mathit{X}=\{x_{i}\}\). We assume that these data contain noise with time-varying variability, which is a frequently encountered aspect in practical scenarios. 
		Similarly to \citet{heskes1996practical}, we model our series as
		\begin{equation}\label{equa: data error term}
			t_{i} = f(\vec{x}_{i}) + \epsilon_{x_{i}},
		\end{equation}
		Where \(\epsilon_{x_{i}}\) is the random time varying noise in the data. And $f(\cdot)$ transforms the observed times series into a quantity of interest denoted by \(t_{i}\). For example, it could take out seasonality and other autocorrelation effects to yield a residual vibration component. As the relationship of $f(\cdot)$ can be complex, we use an LSTM model to approximate it as \(\hat{f}(\vec{x}_{i})\). However, any trained model will always have some modeling error, denoted by $\epsilon_{\hat{f}}$. Thus, Equation (\ref{equa: data error term}) can be written as 
			\begin{equation}\label{equa: two erro terms}
			t_{i} - \hat{f}(\vec{x}_{i}) = (f(\vec{x}_{i}) -\hat{f}(\vec{x}_{i})) +\epsilon_{x_{i}}.
			\end{equation}
		Assuming that the two error terms are independent, the standard deviations can be written as:
			\begin{equation}\label{equa: var of two error terms}
				\sigma_{total}^{2} =\sigma_{\hat{f}}^{2}+\sigma_{\epsilon_{x_{i}}}^{2}.
			\end{equation}
		Where \(\sigma_{total}^{2}\) denotes the variance of two error term, \(\sigma_{\hat{f}}^{2}\) represents the variance of the error term from the model, and \(\sigma_{\epsilon_{x_{i}}}^{2}\) is the time dependent variance of the data noise.
		
		In addition, the dataset $\mathit{X}$ is split into training and test sets. In order to apply the LSTM network, both data sets are restructured by the moving window method with length $w$. That is, the two sets consist of \textit{w}-dimension vectors $\vec{x}_{i}=\left(x_{i}, x_{i+1}, \dotsb, x_{i+w-1}\right)$. Therefore, we have $\mathit{X}_{train} = \{\vec{x}_{p}\}$ and $\mathit{X}_{test} = \{\vec{x}_{q}\}$.
		
		\begin{figure}
			\includegraphics[width=\linewidth]{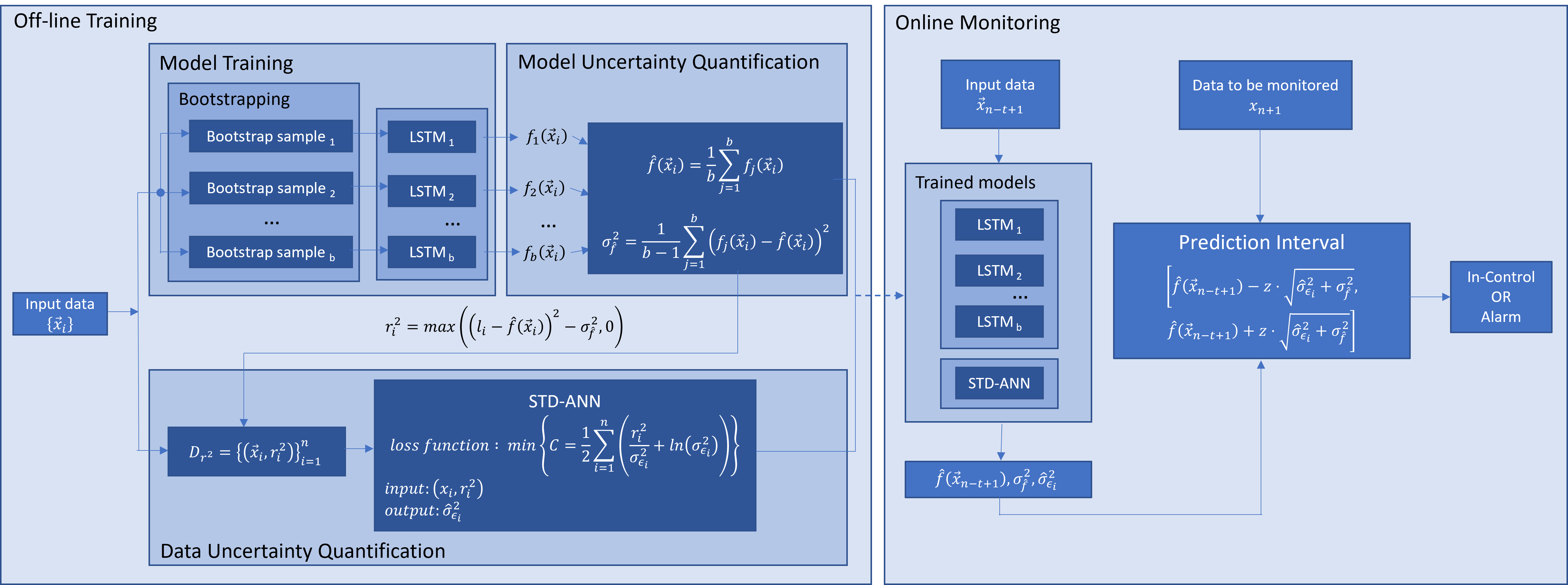}
			\caption{Proposed Model Framework}
			\label{fig:framework}
		\end{figure}
		\subsection{Phase I: Offline Training}
		Our proposed model is illustrated in Figure \ref{fig:framework} and consists of two phases. The first phase is the offline training phase, while the online monitoring is the second phase. In phase I, the prediction model and the variability estimation model are trained. These models are used in Phase II to detect whether the data are in-control. In the offline training, there are three steps: model training, model uncertainty quantification and data uncertainty quantification. 
		%The prediction uncertainty comes from data $\epsilon_{x_{i}}$ and model $\epsilon_{y}$.
		
		The dataset is reorganized to a set of data pairs $\left\{(\vec{x}_{i}, l_{i})\right\}$ to train the LSTM networks, where $\vec{x}_{i}=\left(x_{i}, x_{i+1}, \dotsb, x_{i+w-1}\right)$ is the input and $l_{i}=x_{i+w}$ is the value of the one-time step after $\vec{x}_{i}$ as the label. 
		The bootstrap method is employed to reduce and estimate the epistemic uncertainty. Thus, the data pairs $(\vec{x}_{i}, l_{i})$ in $\mathit{X}_{train}$ are resampled with the resample size \(n\). There are \(b\) bootstrap subsets regarded as \(B_{1}, …, B_{b}\). In each set \(B_{j}\), an LSTM model \(f_{j}\) is trained, as shown in the ``Model Training'' block in upper left corner of Figure \ref{fig:framework}. The training is finished by early-stopping through the model performance on the validation dataset. This validation set comprises the data points in $\mathit{X}_{train}$ but not in \(B_{j}\), also called the out-of-bag (OOB) samples. The early stopping is employed to avoid over-fitting. Over-fitted model can always send false alarms while monitoring, as the model becomes overly sensitive to the specific training sets.
		
		After training \(b\) LSTM models, a less biased prediction $\hat{f}(\vec{x}_{i})$ is made, with regard to \(\vec{x}_{i}\), by averaging all the outputs from \(b\) subsets, i.e.,
			\begin{equation}\label{equa: true f}
				\hat{f}(\vec{x}_{i}) = \dfrac{1}{b}\sum_{j=1}^{b}f_{j}(\vec{x}_{i}).
			\end{equation}
		Next, the variance of the model uncertainty can be estimated, as shown in the ``Model Uncertainty Quantification'' block in Figure \ref{fig:framework}, as:
			\begin{equation}\label{equa: variance of model error}
				\hat{\sigma}_{\hat{f}}^{2} = \dfrac{1}{b-1}\sum_{j=1}^{b}\left(f_{j}(\vec{x}_{i})-\hat{f}(\vec{x}_{i})\right)^{2}.
			\end{equation}
		
		The time varying data noise terms can be estimated after the model uncertainty is given based on Equation (\ref{equa: var of two error terms}). It is obvious that 
		$$
		\sigma_{\epsilon}^{2} \simeq E((t - \hat{f})^{2}) - \sigma_{\hat{f}}^{2}. 
		$$
		Thus, we can calculate the remaining residuals $r_{i}^{2}$ to estimate the variance of the noise by 
		$$
		r_{i}^{2} = \text{max}\left\{\left(l_{i} - \hat{f}(\vec{x}_{i})-\sigma_{\hat{f}_{i}}^{2},\right)\right\},
		$$
		where $r_{i}^{2}$ and the corresponding $\vec{x}_{i}$ compose a set of pairs $D = \left\{\left(r_{i}^{2},\vec{x}_{i}\right)\right\}$. Another ANN \(F(\cdot)\) is trained on \(D\) to estimate the variance of data noise by minimizing the negative component in the log-likelihood function, i.e., the loss function \(C\) is denoted as
			\begin{equation}\label{equa:loss function}
				C = \frac{1}{2}\sum_{i=1}^{n}\left(\frac{r^{2}_i}{\sigma_{\epsilon_{i}}^{2}}+\ln(\sigma_{\epsilon_{i}}^{2})\right).
			\end{equation}
		Thus, the data noise \(\sigma_{\epsilon_{i}}\) can be estimated as \(\hat{\sigma}_{\epsilon_{i}}^{2}=F(\vec{x}_{i})\), like presented in the ``Data Uncertainty Quantification'' block in Figure \ref{fig:framework}.
		
		After the three steps in Phase I, we construct Shewhart-type charts based on the prediction $\hat{f}(\vec{x}_{i})$ for $\vec{x}_{i+w}$ by setting the upper control limit (UCL) and lower control limit (LCL) equal to
			\begin{equation}\label{equa: prediction intervals}
				\begin{aligned}
					UCL_{i+w} = \hat{f}(\vec{x}_{i})+z \cdot s(\vec{x}_{i}),\\
					LCL_{i+w} = \hat{f}(\vec{x}_{i})-z \cdot s(\vec{x}_{i}),
				\end{aligned}
			\end{equation}
		where z is an appropriate quantile from the standard normal distribution, and \( s(\vec{x}_{i})=\sqrt{\hat{\sigma}_{\epsilon_{i}}^{2}+\hat{\sigma}_{\hat{f}}^{2}} \).

		\subsection{Phase II Online Monitoring}
		In the online monitoring phase, without loss of generality, we assume that the \(x_{n+1}\) is the point for monitoring. We then select \(\vec{x}_{n-w+1}=(x_{n-w+1},x_{n-w+2},...,x_{n})\) as the input to the model. The estimated prediction $\hat{f}(\vec{x}_{n-w+1})$ and the variance of two error term $s(x_{i})$ are equal to 
		
		$$\hat{f}(\vec{x}_{n-w+1})=\frac{1}{b}\sum_{j=1}^{b}f_{j}(\vec{x}_{n-w+1}), s(\vec{x}_{n-w+1})=\sqrt{\hat{\sigma}_{\epsilon}^{2}+\hat{\sigma}_{\hat{f}}^{2}},$$
		where, $\hat{\sigma}_{\hat{f}}^{2} = \dfrac{1}{b-1} \sum_{j=1}^{b} \left(f_{b}(\vec{x}_{n-w+1}) - \hat{f}(\vec{x}_{n-w+1}) \right)^{2}, \text{and } \hat{\sigma}_{\epsilon_{n-w+1}}^{2}=F(\vec{x}_{n-w+1}).$ Then, the control limits are set as
		$$ UCL_{n+1} = \hat{f}(\vec{x}_{n-w+1}) + z\cdot s(\vec{x}_{n-w+1}), $$
		$$ LCL_{n+1} = \hat{f}(\vec{x}_{n-w+1}) - z\cdot s(\vec{x}_{n-w+1}). $$
		Thus, \(x_{n+1}\) is in-control if \(x_{n+1}\in [LCL_{n+1},UCL_{n+1}]\), otherwise, it is deemed to be out-of-control.
		
		The computational costs are mainly from the offline training. The cost in online monitoring is only for applying the model to \(\vec{x}_{n+1}\).

	\section{Simulation Study}
		The proposed method is implemented on simulation data to evaluate its effectiveness, and the performances of our method is compared to a benchmark method as well as ablated versions of our proposal method.. The generation of simulation data is presented in Section 4.1. The metrics for evaluating the model performances are introduced in Section 4.2. We present two ablated models alongside a benchmark model as comparisons in Section 4.3. The comparison model setups and results with analysis are discussed in Section 4.4.
		\subsection{Experiment Description}
		We wish to simulate data with autocorrelation and time-varying variability. 
		We believe that these are two common aspects observed in real applications.
		Hence, the simulation data are generated by AR(1)-GARCH(1,1) model, i.e., the data linearly depends on the previous values and on a time-varying error term. The formulation is as follows; 
		\begin{equation}\label{equa:ARGRACH}
			x_{t}=\phi \cdot x_{t-1}+\epsilon_{t}
		\end{equation}
		where $\epsilon_{t} = \sigma \cdot z_{t}$, $z_{t} \sim N(0,1)$, and $\sigma^{2}_{t} =\alpha_{0} + \alpha_{1} \cdot \epsilon^{2}_{t-1}+ \beta \cdot \sigma^{2}_{t-1}$. 
		There are $T=500$ points generated by (\ref{equa:ARGRACH}) with a random seed.
		A mean shift $\delta$ is set equal to $(0,0.25,0.50,0.75,1.0,1.5,2.0)$ from $\tau=401$. This means that the first 400 data points are in-control and the next 100 data points are out-of-control with a mean shift. When the shift is 0, all points are in-control. 1000 random seeds are selected from a range of 20,000 to 120,000 in increments of 100. This means the experiments are repeated $r=1000$ times on data generated using a different seed. The GARCH parameters are set to be constant in all experiments, while the AR parameter $\phi$ used in the simulation is set to at \(0.1,0.5\) and \(0.9\). Thus, there are 21000 sets of simulation data generated. 
		\subsection{Performance Metrics}
			There are four performance metrics used: False Alarm Probability (FAP), Conditional Expected Delay (CED), Detection Rate (DR), and Detection Recall Rate (DRR). 
			Denote the time of change by \(\tau\), for our simulation study \(\tau=401\). And denoted by \(t_{A}\) the time of first alarm.
			
			The empirical FAP is used to evaluate the in-control performance of the selected models, denoted by $P(t_{A}<\tau)$. 
			The FAP should ideally be equal to a pre-selected rate \(\alpha\) used to set \(z\) in Equation (\ref{equa: prediction intervals}). The quantile $z$ is directly controlling the FAP, where a larger $z$ makes the FAP smaller. The FAP is also affected by the hyper-parameters, as they influence the learning performances of the neural network models. Thus, a poor learning performance will lead to a poor FAP. The window size also plays an essential role in controlling FAP. A relatively smaller window size indicates a higher FAP.
			
			The CED measures conditional time steps between the time of change $t=\tau$ and the time of an alarm $t_{A}$. From \citet{frisen2009optimal}, it is denoted by 
				\begin{equation}\label{equa: CED}
					CED(t) = E\left[t_{A}-\tau|t_{A}\geq\tau\right].
				\end{equation}
			A smaller CED indicates a quicker detection of the change. 
			
			The DR shows the ability whether a model can detect the change, and it is denoted by
				\begin{equation} \label{equa: DR}
					DR = E\left[I\left[\tau \leq t_{A}\leq T\right]\right]
				\end{equation}
			where $I$ is the indicator function denoting if a signal is observed or not, and $T$ is the size of the given dataset.
			
			In this simulation, the control charts will not be reset when it triggers an alarm. Thus, we can observe multiple alarms in the \(T\) data points and the more alarms signalled, the more accurate the charts can detect changes. Based on this, we include another metric that can measure this accuracy, the Recall Rate, which is the fraction of the points after the change detected as alarms, denoted by
			 	\begin{equation}\label{equa: Recall}
			 		Recall = \frac{n}{T-\tau}
			 	\end{equation}
		 	where $n$ is the number of triggered alarms and $T-\tau$ is the total number of points with a mean shift. In our simulation, $T-\tau=100$. A larger recall rate indicates better detection performance.
		
		\subsection{Comparison Models}
			To evaluate the performance of our proposed method, we compare its performance with a model proposed by \citet{chen2019deep}. The reason why this method is chosen is that the structure of this model is comparable to the proposed model. In addition, we execute an ablation study, including two ablated models for comparisons. These benchmark methods are applied to showcase the importance of each part of the proposed method. The differences between our model and these three models are presented in Table \ref{table:comparison}.
			
			\begin{table}[h!]
				\tbl{Model Comparison}
				{\begin{tabular}{c c c c c}\toprule
					\multirow{2}{*}{}& Proposed  & Ablated  & Ablated  & RNN Model \\
					 &   Model &  Model A& Model B &  \citep{chen2019deep}\\
					\hline
					Neural Network Model &LSTM&LSTM&LSTM&RNN\\
					\hline
					Bootstrap Method & \usym{2714} &\usym{2714} &\usym{2715} &\usym{2715}\\
					\hline
					Uncertainty Quantification & \usym{2714} &\usym{2715} &\usym{2715} &\usym{2715}\\
					\bottomrule
				\end{tabular}}
%				\caption{Model Comparison}
				\label{table:comparison}
			\end{table}
			\subsubsection{RNN-Residual Charts}
				This method proposed by \citet{chen2019deep} uses RNN to fit the time series model. The RNN is well-trained when the prediction errors meet the predetermined requirement. The distribution of residuals between prediction and sample points is then estimated. The residual charts are built by the estimated variance and mean. This estimated mean shall be small and close to zero. This method has a comparable structure to our model. The main difference between the two methods is the approach to constructing the control chart. This method determines the control limit directly based on the distribution of residuals, while our method gives more precise control limits by quantifying the time varying data noise and model error separately. Since there is no open-access code of the method, we wrote the code ourselves based on their conception. 
			\subsubsection{Ablation Study}
				The idea of an ablation study in machine learning is used to understand more about a model's behaviour. In our comparison, we design two ablated models. The first one, named as Ablated A, ablates the part of estimating the time varying noise, i.e., the ANN used to estimate the data noise (the ``Data Uncertainty Quantification'' block in Figure \ref{fig:framework}). Thus, the control charts are built by the distribution of residuals between the less biased predictions and labels. The second model is called Ablated B, which ablates the part of the prediction interval construction, i.e., both bootstrap and ANN (the ``Bootstrapping'' and ``Data Uncertainty Quantification'' blocks in Figure \ref{fig:framework}). This means that the LSTM model is used to predict the value at the next time step, and the residuals between the predicted value and the observed value are calculated. Based on the distribution of these residuals, the residual chart is built. Hence, the Ablated model B is comparable to the RNN-residual charts but uses an LSTM neural network.
		\subsection{Simulation Results}
			Given 500 generated simulation data points, the first 350 of 400 in-control points are used for training, while 50 in-control and 100 out-of-control points are used for testing. The simulation data are re-structured by the moving window method for the LSTM and RNN. 
		
			There are two simulation experiments made, and they are set up as follows. The first experiment is used to decide the hyper-parameters, while the second is set to compare the performances of the three models. In the first simulation, models are implemented on the datasets where $\delta=0$. The in-control performances of the three methods are measured by the FAP. Hyper-parameters of the three 	models are settled when the FAP is close to 0.02. The FAP of the three methods is shown in Table \ref{table: FAP}. The hyper-parameters of the proposed model are shown in Table \ref{table:parameters}.
		
			\begin{table}[h]
				\tbl{FAP of four models}
				{\begin{tabular}{c c c c c} \toprule
					AR parameters& LSTM & Ablated A& Ablated B & RNN\\
					\hline
					$\phi = 0.1$ &0.0185&0.0177&0.0175 & 0.0178\\
					$\phi = 0.5$ &0.0186&0.0195&0.0189 & 0.0187\\
					$\phi = 0.9$ &0.0174&0.0237&0.0211 & 0.0204\\
					\bottomrule
				\end{tabular}}
				\label{table: FAP}
			\end{table}

			In the second simulation, three models with the aforementioned parameters are implemented in the simulation datasets where $\delta>0$. The DR, CED and recall rate are calculated to compare their performances.
		\begin{table}[h]
			\tbl{{Hyper-parameters of the proposed model}}
			{\begin{tabular}{c c c c} 
				\hline
				 Window Size & Learning Rate & Epoch & Batch Size\\ [0.5ex] 
				 \hline
				 5 & 0.01 & 300 & 32 \\
				\hline
			\end{tabular}}
%			\caption{Hyper-parameters of the proposed model}
			\label{table:parameters}
		\end{table}

\begin{figure}[h!]
			\subfigure[$\phi = 0.1$ \label{fig:DR_0.1}]{\resizebox*{0.33\linewidth}{!}{%
				\includegraphics{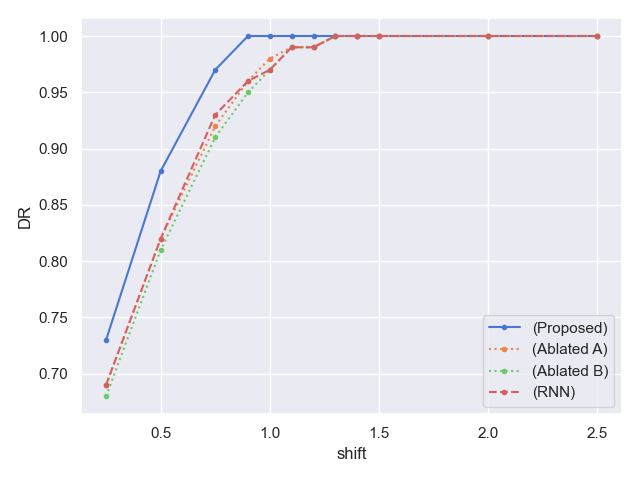}}}
			\subfigure[$\phi = 0.3$ \label{fig:DR_0.5}]{\resizebox*{0.33\linewidth}{!}{%
				\includegraphics{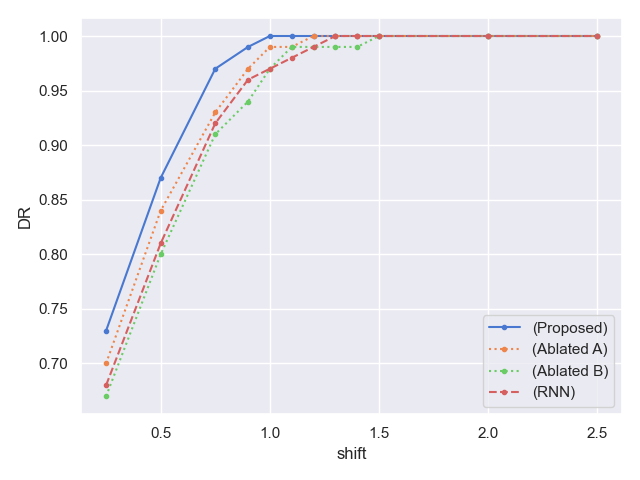}}}
			\subfigure[$\phi = 0.9$ \label{fig:DR_0.9}]{\resizebox*{0.33\linewidth}{!}{%
				\includegraphics{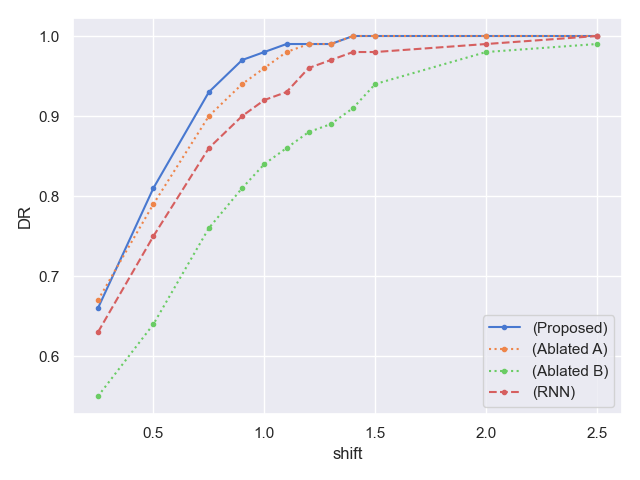}}}
			\caption{DR Comparison of four models}
			\label{fig:DR}
		\end{figure}

\begin{figure}[h!]
			\subfigure[$\phi = 0.1$ \label{fig:CED_0.1}]{\resizebox*{0.33\linewidth}{!}{%
				\includegraphics{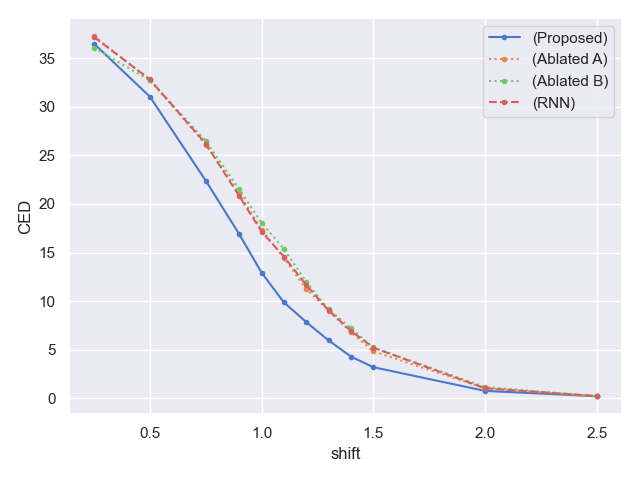}}}
			\subfigure[$\phi = 0.3$ \label{fig:CED_0.5}]{\resizebox*{0.33\linewidth}{!}{%
				\includegraphics{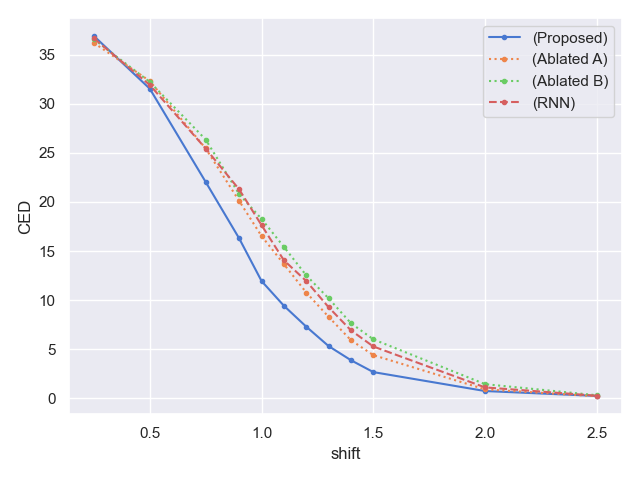}}}
			\subfigure[$\phi = 0.9$ \label{fig:CED_0.9}]{\resizebox*{0.33\linewidth}{!}{%
				\includegraphics{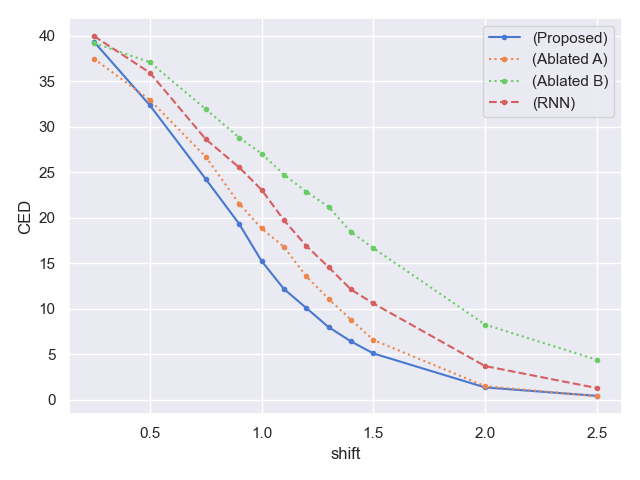}}}
			\caption{CED Comparison of four models}
			\label{fig:CED}
		\end{figure}

\begin{figure}[!]
			\subfigure[$\phi = 0.1$ \label{fig:Recall_0.1}]{\resizebox*{0.33\linewidth}{!}{%
				\includegraphics{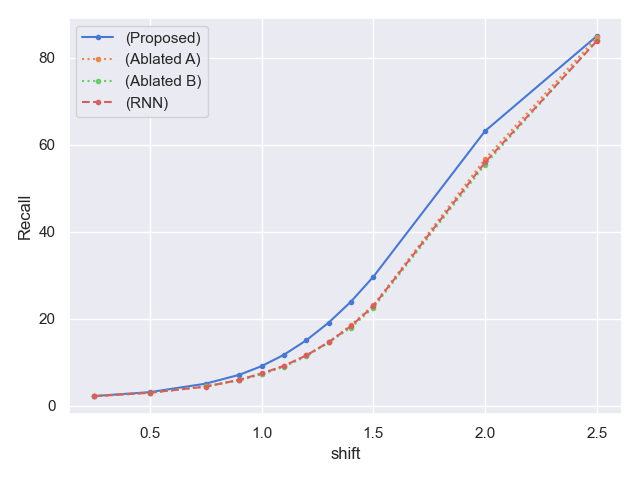}}}
			\subfigure[$\phi = 0.3$ \label{fig:Recall_0.5}]{\resizebox*{0.33\linewidth}{!}{%
				\includegraphics{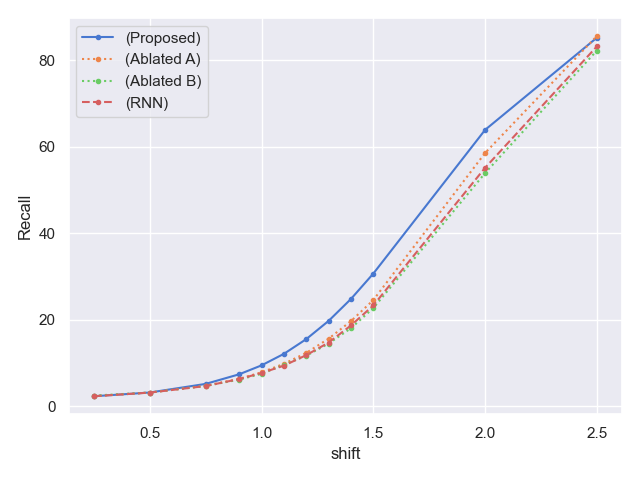}}}
			\subfigure[$\phi = 0.9$ \label{fig:Recall_0.9}]{\resizebox*{0.33\linewidth}{!}{%
				\includegraphics{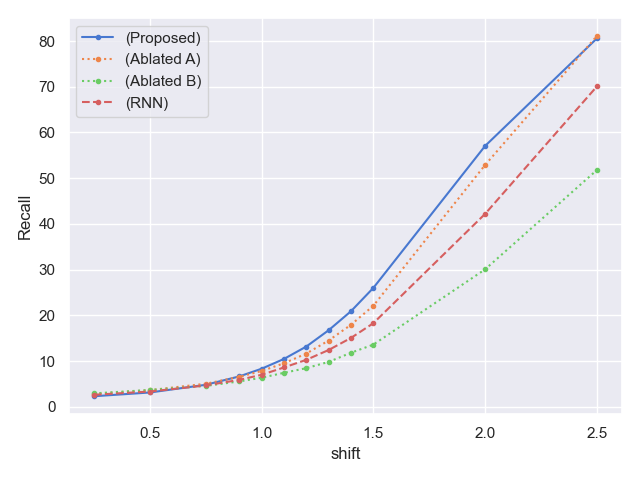}}}
			\caption{Recall Comparison of four models}
			\label{fig:Recall}
		\end{figure}

		Table \ref{table:results} presents the DR, CED and Recall of three models on various AR(1)-GARCH(1) processes with different mean shifts, and Figures \ref{fig:DR} to \ref{fig:Recall} illustrate the performances of the four models. The proposed method outperforms the other methods in most instances. There are only slight differences among the four models' performances on processes with a tiny mean shift $(\delta=0.25)$ or a large mean shift $(\delta \geq 2)$ in most scenarios.
		
		Firstly, the proposed model has the highest DR in all scenarios. The DR of all methods can reach $95\%$ when the mean shift is not small $(\delta \geq 1.0)$, except the ablated method on a highly auto-correlated process with $ \phi = 0.9$, as listed in Table \ref{table:results} and illustrated in Figure \ref{fig:DR}.

		Secondly, as presented in Figure \ref{fig:CED}, all four methods can detect significant changes $(\delta \geq 1.5)$ within 5 time steps, if \(\phi=0.1\) or \(0.5\). The ability to detect shifts quickly deteriorates for strongly auto-correlated processes \((\phi=0.9)\) for the ablated B and RNN method.
		Except for situation when the mean shifts is tiny$(\delta = 0.25)$, the proposed model can always manage to detect signals earlier than others.
		Thirdly, as displayed in Figure \ref{fig:Recall}, there are always more points that the proposed model can detect. However, all methods capture only a few data with a mean shift when the $\delta$ is not larger than $1.0$. More specifically, when $\delta \leq 0.75$, the recall rate of the four methods are relatively close to each other, as their difference is less than 1. Meanwhile, all methods can only capture less than $10\%$ signals when $\delta$ is small ($\delta \leq 1$).
		
		Fourthly, the results of the ablated model B and RNN residual charts show that the choice of RNN or LSTM does not show a huge difference, when $\phi \leq 0.5$. But RNN charts work better than ablated B, when $\phi = 0.9$. This is expected as RNN theoretically should outperform an LSTM when the long-term memory is required (i.e., strongly auto-correlated processes). The comparison results between the two ablated models provide the evidence to confirm that obtaining less biased predictions by bootstrapping is essential for better anomaly detection, especially, when $\phi=0.9$. The uncertainty quantification is also essential, as the proposed model outperforms the ablated model A in most cases.
		
		Finally, all these conclusions have been replicated on an second simulation study with synthetic data generated using a new set of random seeds. More details can be found in the Appendix.
		
		\begin{sidewaystable}[!]
			\tbl{Performance comparison of our proposed method with two ablated models and a benchmark}{
				\begin{tabular}{c c c c c c c c c c c c c c c c }
					\toprule
					AR & Mean &&& DR && &&& CED &&  &&& Recall &\\
					\cline{3-6}\cline{8-11}\cline{13-16}
					Parameters &  Shift & LSTM  & Ablated A & Ablated B& RNN & & LSTM & Ablated A & Ablated B & RNN & & LSTM & Ablated A & Ablated B & RNN\\
					\midrule
					\multirow{6}{*}{$\phi=0.1$}
					& 0.25 &\textbf{0.73}& 0.69 & 0.68& 0.69 &&36.50&37.28&\textbf{36.07} & 37.19&& 2.28& 2.26&  \textbf{2.30}&2.24\\
					& 0.50 &\textbf{0.88}& 0.82 & 0.81& 0.82 &&\textbf{31.05}&32.79&32.71 & 32.81&& \textbf{3.18}& 3.04&  3.07&3.04\\
					& 0.75 &\textbf{0.97}& 0.92 & 0.91& 0.93 &&\textbf{22.41}&26.10&26.51 & 26.21&& \textbf{5.13}& 4.48&  4.50&4.45\\
					& 1.00 &\textbf{1.00}& 0.98 & 0.97& 0.97 &&\textbf{12.94}&17.34&18.07 & 17.16&& \textbf{9.15}& 7.42&  7.28&7.47\\
					& 1.50 &1.00& 1.00& 1.00 & 1.00 &&  \textbf{3.20}&  4.83&   5.23 &   5.21&&\textbf{29.69}&23.13& 22.60&22.94\\
					& 2.00 &1.00& 1.00& 1.00 & 1.00 &&  \textbf{0.76}&  1.05&   1.16 &   1.03&&\textbf{63.19}&56.77& 55.48&55.99\\
					\midrule
					\multirow{6}{*}{$\phi=0.5$}
					&0.25&\textbf{0.73}&0.70&0.67&0.68 &&36.91&\textbf{36.21}&36.58& 36.67&& 2.35& 2.35& \textbf{2.50} &2.36\\
					&0.50&\textbf{0.87}&0.84&0.80&0.81 &&\textbf{31.54}&32.32&32.23& 31.92&& \textbf{3.21}& 3.20& \textbf{3.21}& 3.19\\
					&0.75&\textbf{0.97}&0.93&0.91&0.92 &&\textbf{22.04}&25.45&26.36& 25.44&& \textbf{5.20}& 4.78& 4.72&4.71\\
					&1.00&\textbf{1.00}&0.99&0.97&0.97 &&\textbf{11.94}&16.52&18.27& 17.63&& \textbf{9.48}& 7.88& 7.53&7.72\\
					&1.50&1.00&1.00&1.00&1.00 &&  \textbf{2.66}&   4.39& 6.00&    5.27&&\textbf{30.68}&24.57&22.71&23.32\\
					&2.00&1.00&1.00&1.00&1.00 && \textbf{0.73}&  0.90& 1.43&    1.10&&\textbf{63.87}&58.53&53.89&55.11\\
					\midrule
					\multirow{6}{*}{$\phi=0.9$}
					&0.25&\textbf{0.66}&0.67&0.55&0.63 &&39.37&\textbf{37.50}&39.23& 39.99&& 2.31& 2.82& \textbf{2.93}&2.63\\
					&0.50&\textbf{0.81}&0.79&0.64&0.75 &&\textbf{32.38}&32.96&37.11& 35.96&& 3.13& 3.64& \textbf{3.68}&3.34\\
					&0.75&\textbf{0.93}&0.90&0.76&0.86&&\textbf{24.27}&26.72&31.96& 28.65&& 4.81& \textbf{5.04}& 4.56&4.67\\
					&1.00&\textbf{0.98}&0.96&0.84&0.92 &&\textbf{15.25}&18.83&27.06& 23.08&& \textbf{8.32}& 7.83& 6.36&7.00\\
					&1.50&\textbf{1.00}&\textbf{1.00}&0.94&0.98 &&  \textbf{5.08}&  6.56&16.67& 10.58&&\textbf{26.00}&22.09&13.60&18.28\\
					&2.00&\textbf{1.00}&\textbf{1.00}&0.98&0.99 &&  \textbf{1.34}&  1.48&  8.25&   3.69&&\textbf{57.05}&52.89&30.06&42.19\\
					\bottomrule
			\end{tabular}}
			\label{table:results}
		\end{sidewaystable}

	\section{Case Study}
	In this section, the performance of the proposed method is validated by implementing it on two datasets with time-varying variability: vibration and energy consumption data of escalators in Hong Kong subway stations. In the first case, the variability of the data is time-varying because of component degradation, while seasonality influences the data variability in the energy case. Readers are referred to \citet{zwetsloot2023remaining} for more details of this case study.
	
		\subsection{Vibration}
		The proposed method is applied on an escalator vibration dataset to monitor the operating health status of the escalator. There are eight data streams from  sensors attached to the motor, gearbox, tension carriage and main drive. Data are collected three times a day during the escalator's working period. In this case study, a whole year of data, from January 2021 to January 2022, from a single vibration sensor of an escalator, are selected.

\begin{figure}[!]
			\includegraphics[width=\linewidth]{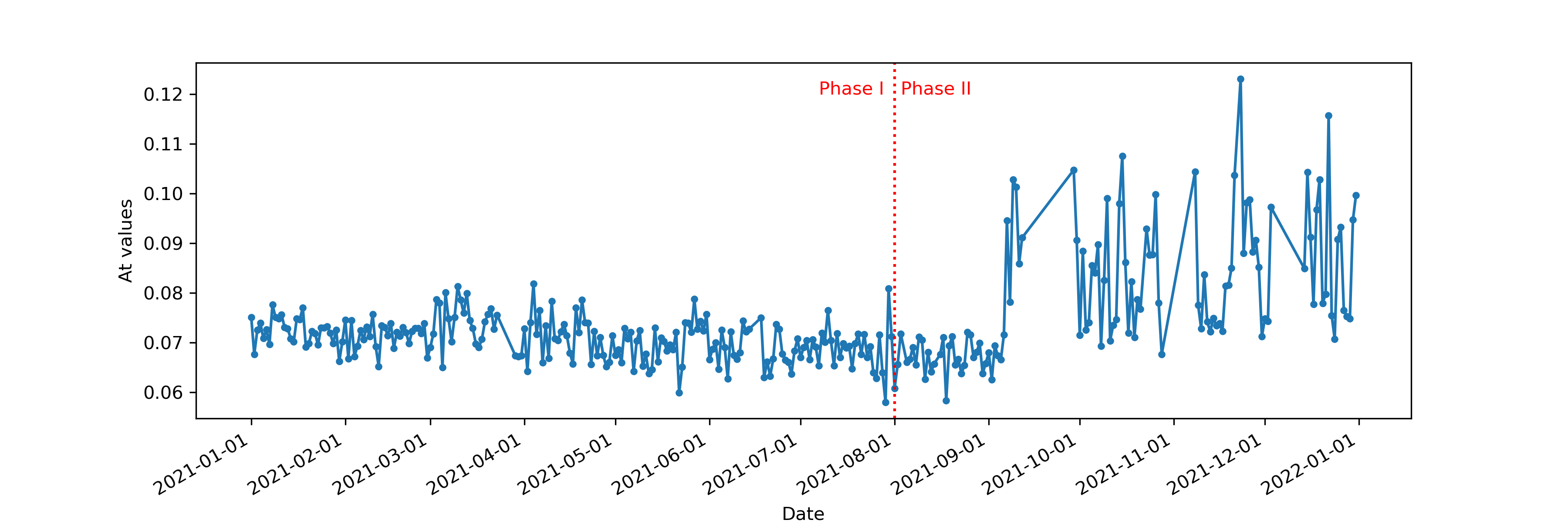}
			\caption{Raw $A_{t}$ values in both Phases}
			\label{fig:AllAt}
		\end{figure}

		The raw vibration data are transferred to the frequency domain by fast Fourier transformation and averaged on each frequency. After that, the averaged vibration spectrum is summarized into $A_{t}$ values, denoted by
			$$ A_{t} = \sqrt{\frac{n^{2}_{1}+n^{2}_{2}+...+n^{2}_{n}}{1.5}} , $$
		where $n_{i}, i=1,2,..., $ is the amplitude value of the corresponding frequency. This is a standard transformation used to compare vibration levels to ISO standards. Figure \ref{fig:AllAt} presents the processed sensor data in the selected period. The x-axis shows the timeline, while the $A_{t}$ values are on the y-axis. Data from January to early August 2021 are regarded as phase I and used for training. The phase II data to be monitored are from August to the end of 2021, and are the testing dataset.
		
		\begin{figure}[!]
			\includegraphics[width=\linewidth]{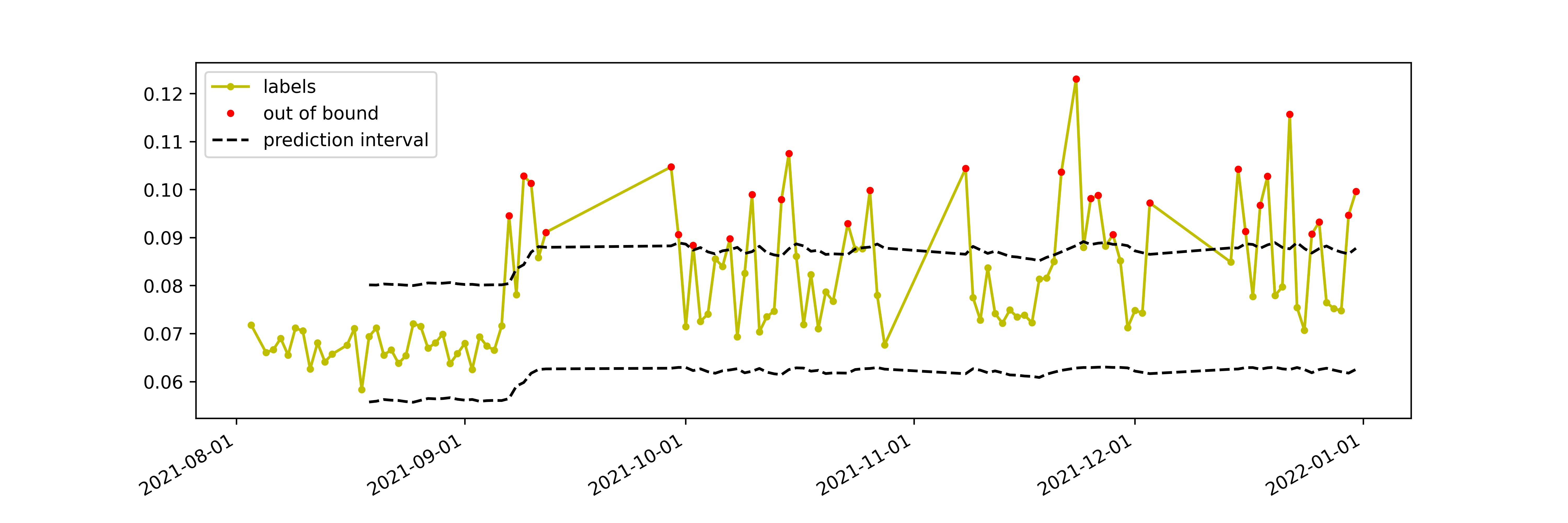}
			\caption{Results on Vibration Test Dataset}
			\label{fig:Test_vib}
		\end{figure}
	
		Preventive maintenance to escalators is performed every two weeks, so we set the length of the moving time window as 14 days. This also means that the passenger pattern repeats twice as it shows a weekly (and daily) pattern. Both training and test datasets are restructured by the moving window. 
		
		For implementation of our method we first train our model on the Phase I data and next run it on the test data which are visualized in Figure \ref{fig:Test_vib}. The proposed charts triggers 29 alarms in total. The first alarm signal is on September 7, which we relate to a preventive maintenance event on the night of September 6. Since then, the escalator operating status seems to have changed.

		\subsection{Energy}
		\begin{figure}[h!]
			\subfigure[Raw Energy Data \label{fig:allenergy}]{%
				\resizebox*{\linewidth}{!}{\includegraphics{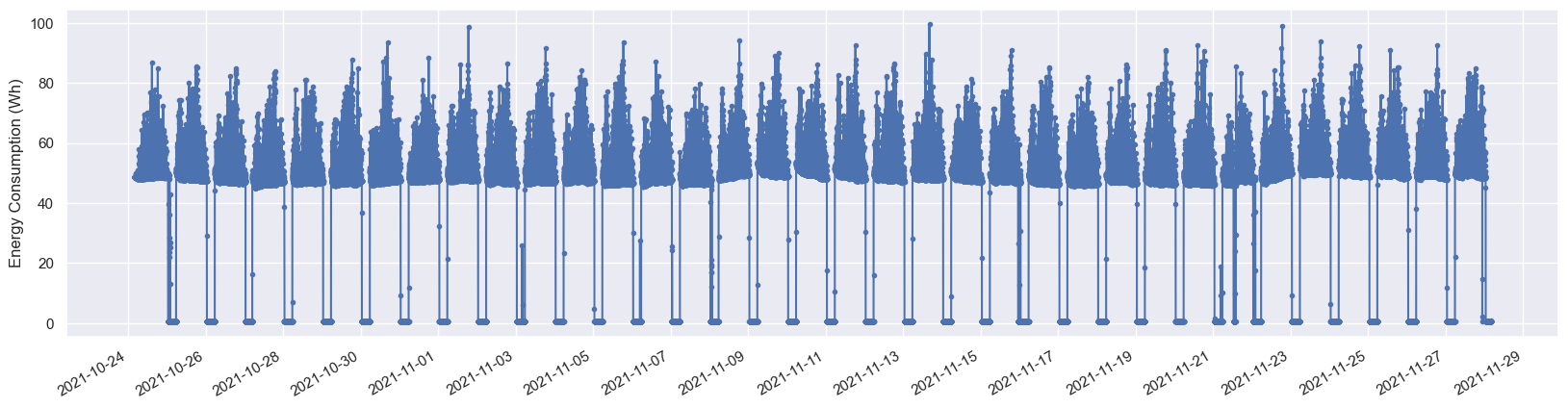}}}
			\subfigure[Processed Energy Data \label{fig:ProcessedEnergy}]{%
				\resizebox*{\linewidth}{!}{\includegraphics[width=\linewidth]{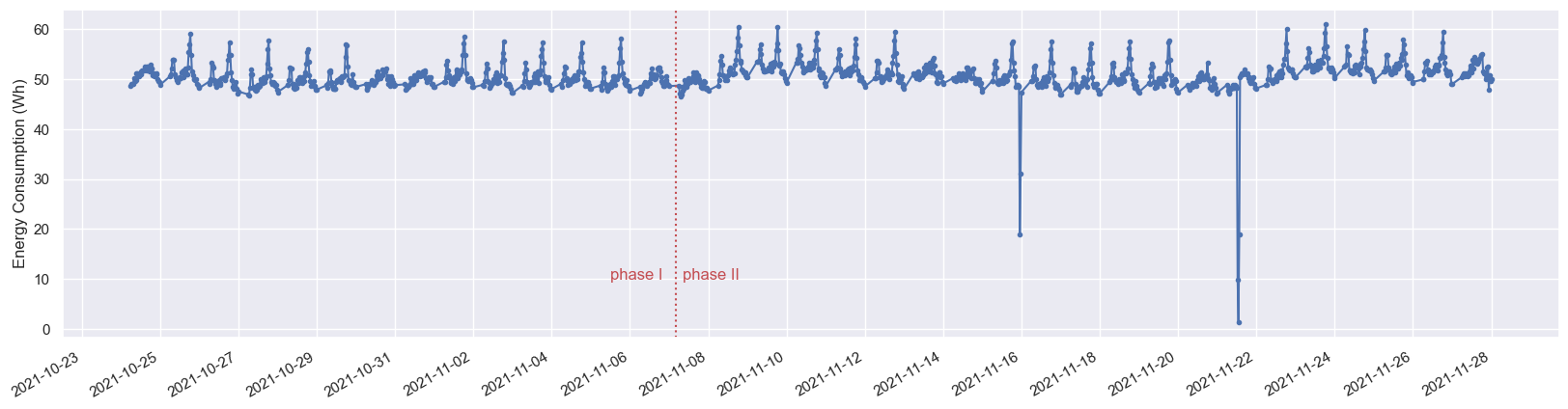}}}

			\caption{Energy Case Study Data}
			\label{fig:EnergyCaseStudy}
		\end{figure}

		On the energy dataset, the proposed model is used to monitor the energy consumption of an upward escalator. Monitoring energy enables real-time detection of unexpected shutdowns. The data are collected from an  energy meter every minute. Thus, there are 1,440 points every 24 hours. As the escalators in Hong Kong subway stations work from about 5:30 am to 12:30 am, service days are defined from 4 am to the following day 4 am. The dates mentioned in this case study are service dates rather than calendar dates. In this case, energy consumption of an escalator in five weeks from October 24 2022 to November 28 2022 are selected, as shown in Figure \ref{fig:allenergy}. 
	
		\begin{figure}[h!]
			\subfigure[Processed Data with Weekly Average \label{fig:ProcessedWithMA}]{%
			\resizebox*{\linewidth}{!}{\includegraphics[width=\linewidth]{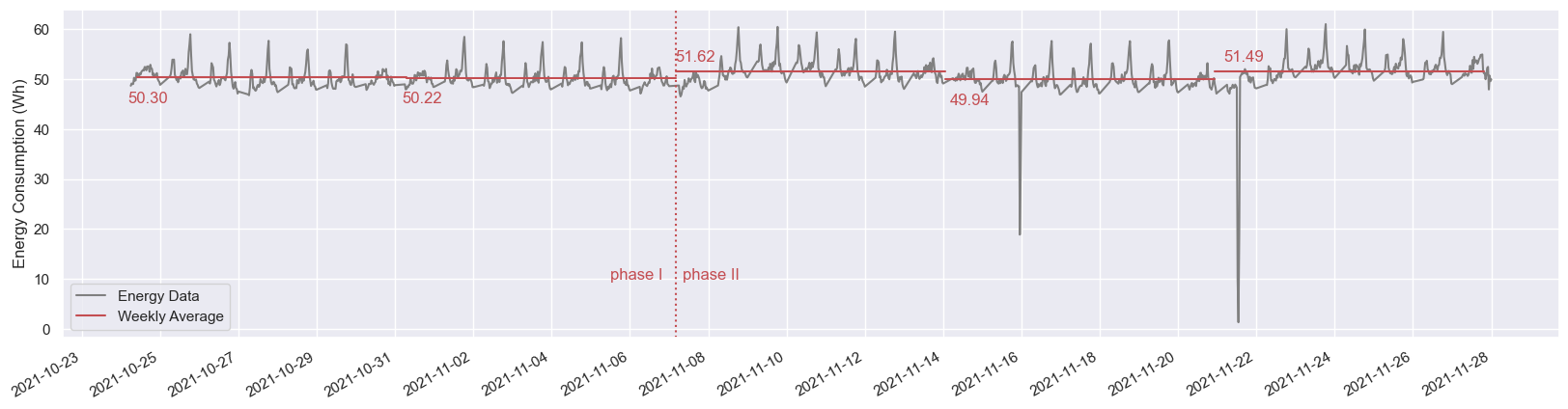}}}
			\subfigure[Moving Range of Processed Data \label{fig:ProcessedOnlyMR}]{*
			\resizebox*{\linewidth}{!}{\includegraphics[width=\linewidth]{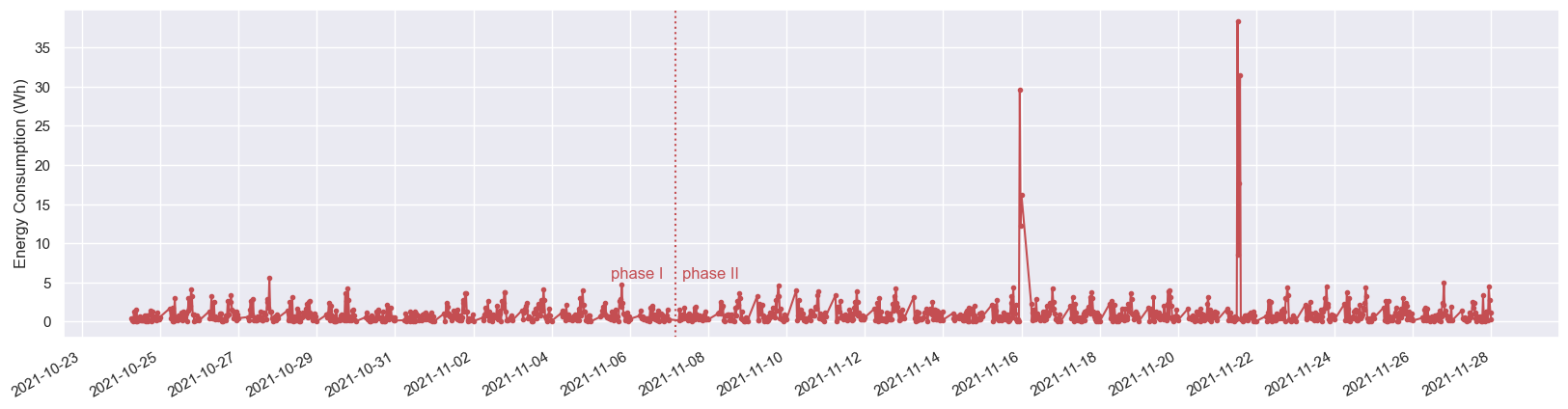}}}
			\caption{Processed Data with Weekly Average and Moving Range}
			\label{fig:ProcessedWithMR&WA}
		\end{figure}

		The minute-wise raw data are averaged every half hour, which scales down the size from 1,440 to 48 points a day. The data collected during the non-working period, from 12:30 am to 5:30 am, are also filtered out, as our focus is on the energy consumption during normal operating time. Therefore, 37 data points for each day are included. The processed data are plotted in Figure \ref{fig:ProcessedEnergy}. The data of the first two weeks, from October 24 to November 7, are used to train the proposed model. The last three weeks' data are used for monitoring. Similarly to Section 5.1, both training and test dataset are restructured by the moving window approach for LSTM model application. And the length of the moving window is set to 37, which is the length of daily energy data and takes the daily pattern into consideration.

		\begin{figure}[h!]
			\includegraphics[width=\linewidth]{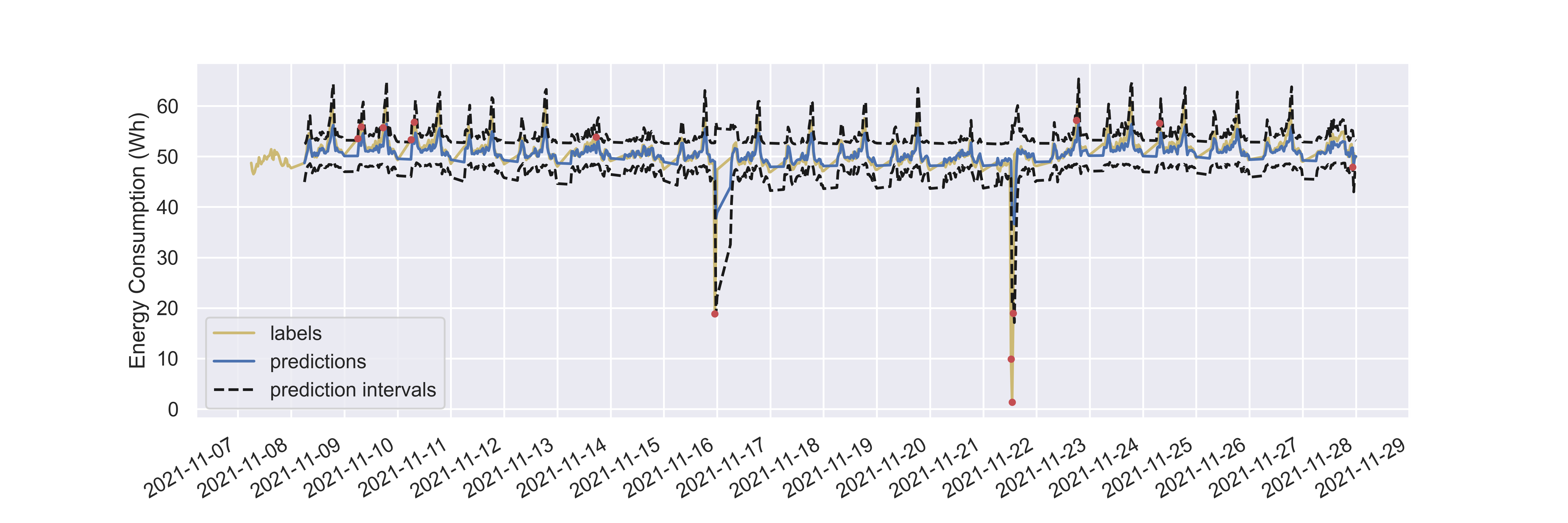}
			\caption{Results on Energy Test Dataset}
			\label{fig:Test_energy}
 		\end{figure}
		From the processed data in Figure \ref{fig:ProcessedEnergy}, it is obvious there are two different energy consumption patterns on weekdays and weekends, which indicates the processed data present two seasonal components. In addition, there are two drops on November 15 and 21, respectively, which indicate two unexpected shutdowns during operation. The five weekly averages are plotted in Figure \ref{fig:ProcessedWithMR&WA} as five horizontal lines. They show that the first two and the fourth weeks have relatively constant averages while there are shifts ($\delta>0$) in the third and final week. The moving ranges are also computed to study the variability of all processed data over time, presented in Figure \ref{fig:ProcessedWithMR&WA}. Except the two extremely large jumps in the data, the moving ranges show the similar seasonality as the energy data itself. That is, the variability is time-varying. Thus, the proposed method is expected to fit well on this dataset.

		After the model is trained, the monitor results are depicted in Figure \ref{fig:Test_energy}. There are thirteen signals triggered by various reasons. The first six alarms are caused by the mean shift, and the four signals on November 15 and 21 reflects two unexpected shutdown. In the last week, there are three alarms raised, two of which are caused by another mean shift, while the last of which we deem as a false alarm.
		
	\section{Conclusion}
	This paper proposed an effective approach to predictively monitoring of sensor data based on LSTM networks.
	The time series sensor data are fed into bootstrapped LSTM models and an ANN. The outputs of LSTM models are used to estimate a less biased prediction and the epistemic uncertainty, while the ANN outputs the estimated variances of sensor data noise. Finally, a Shewhart-type control chart is constructed on the predictions and these two variances. By applying the LSTM model, this work does not rely on the labelled data in the training. And the uncertainty quantification modules allow the method to deal with the time-varying variability. 
	
	The proposed model is compared with other methods in simulations using CED, DR, and recall rate. The comparison results show that quantifying the variances separately makes the proposed model significantly outperform the other models. And the escalator case study tests the performance of the proposed model. The results confirm the effectiveness and usability of the model. In the future, this model can be developed to monitor multivariate data.

	\appendix
	\section{Validation simulation study}
	In this appendix, a second simulation experiment is conducted for validating the performances of our proposed model.
	In this experiment, the random seeds are selected in a different way.
	We set the random seeds from 200 to 3200 in steps of 3. Apart from the random seeds all other simulation settings have been kept identical to those described in Section 4.1. 
	Under this scenario, Table \ref{table: FAP_ref} lists the FAP of the four models. And the results of CED, DR and recall rate are listed in Table \ref{table:resultsCompRef}, which validate the out-performance of the proposed model. 
	
	The differences between the two simulations are compared in absolute percentages, which are presented in brackets in Table \ref{table:resultsCompRef}. The changes of most results are less than $5\%$. When the AR parameter $\phi=0.9$ and the mean shift $\delta=2.0$, the CED of the ablated model B has $13.3\%$ absolute percentage difference on two simulations, which is the largest change. Overall it can be concluded that the results are only slightly influenced by the choice of random seed.
	\begin{table}[h]
		\tbl{FAP of all methods on different random seeds selection}
		{\begin{tabular}{c c c c c}
			\toprule
			AR parameters&LSTM   &Ablated A & Ablated B & RNN    \\
			\midrule
			$\phi = 0.1$ &0.0182 & 0.0177   & 0.0174    & 0.0179 \\
			$\phi = 0.5$ &0.0181 & 0.0195   & 0.0198    & 0.0192 \\
			$\phi = 0.9$ &0.0173 & 0.0238   & 0.0220    & 0.0201 \\
			\bottomrule
		\end{tabular}}
		\label{table: FAP_ref}
	\end{table}
		
	\begin{sidewaystable}[!]
	\tbl{Results of another simulation with different random seeds selection}{
		\begin{tabular}{c c |c c c c c c c c c c c c c c }
		\toprule
		AR & Mean &&& DR && &&& CED &&  &&& Recall &\\
		\cline{3-6}\cline{8-11}\cline{13-16}
		Parameters &  Shift & LSTM  & Ablated A & Ablated B& RNN & & LSTM & Ablated A & Ablated B & RNN & & LSTM & Ablated A & Ablated B & RNN\\
		\midrule
		\multirow{6}{*}{$\phi=0.1$}
		&0.25&\textbf{0.74} (1.37)&0.70 (1.45)&0.68 (0.00)&0.70 (1.45)&&\textbf{37.84} (3.67)&38.94 (4.45)&39.03 (8.21)&38.76 (4.22)&&\textbf{ 2.17} (4.82)& \textbf{2.17} (3.98)& 2.15 (6.52)&   2.12 (5.36)\\
		&0.50&\textbf{0.89} (1.14)&0.85 (3.66)&0.83 (2.47)&0.85 (3.66)&&\textbf{30.67} (1.22)&31.62 (3.57)&32.58 (0.40)&31.37 (4.39)&&\textbf{ 3.15} (0.94)& 2.97 (2.30)&   2.93 (4.56)& 3.00 (1.32) \\
		&0.75&\textbf{0.96} (1.03)&0.95 (3.26)&0.94 (3.30)&0.94 (1.08)&&\textbf{20.83} (7.05)&24.81 (4.94)&24.81 (6.41)&24.55 (6.33)&&\textbf{ 5.17} (0.78)& 4.46 (0.45)&   4.39 (2.44)& 4.47 (0.45) \\
		&1.00&\textbf{1.00} (0.00)&0.98 (0.00)&0.97 (0.00)&0.98 (1.03)&&\textbf{12.52} (3.25)&16.51 (4.79)&17.23 (4.65)&16.37 (4.60)&&\textbf{ 9.33} (1.97)& 7.60 (2.43)&   7.38 (1.37)& 7.59 (1.61) \\
		&1.50&        1.00  (0.00)&1.00 (0.00)&1.00 (0.00)&1.00 (0.00)&&\textbf{ 2.99} (6.56)& 4.88 (1.04)& 5.11 (2.29)& 4.91 (5.76)&&\textbf{30.35} (2.22)&23.83 (3.03)&  23.12 (2.30)&23.65 (3.10) \\
		&2.00&        1.00  (0.00)&1.00 (0.00)&1.00 (0.00)&1.00 (0.00)&&\textbf{ 0.73} (3.95)& 0.98 (6.67)& 1.05 (9.48)& 1.03 (0.00)&&\textbf{63.72} (0.84)&57.45 (1.20)&  56.20 (1.30)&56.64 (1.16) \\
		\midrule
		\multirow{6}{*}{$\phi=0.5$}
		&0.25 &  \textbf{0.76} (4.11)&  0.72 (2.86)&  0.68 (1.49)&  0.70 (2.94)&&  \textbf{37.84} (2.52)&  37.95 (4.81)&  38.16 (4.32)&  38.64 (5.37)&&            2.17 (7.66)&   2.29 (2.55)&  \textbf{ 2.38} (4.80)&   2.31 (2.12)\\
		&0.50 &  \textbf{0.90} (3.45)&  0.87 (3.57)&  0.82 (2.50)&  0.84 (3.70)&&  \textbf{30.20} (4.25)&  31.18 (3.53)&  31.85 (1.18)&  31.99 (0.22)&&            3.15 (1.87)&   3.16 (1.25)&  \textbf{ 3.19} (0.62)&   3.04 (4.70)\\
		&0.75 &  \textbf{0.97} (0.00)&  0.95 (2.15)&  0.92 (1.10)&  0.94 (2.17)&&  \textbf{20.69} (6.13)&  23.59 (7.31)&  25.30 (4.02)&  24.93 (2.00)&&  \textbf{ 5.25} (0.96)&   4.77 (0.21)&            4.85 (2.75)&   4.68 (0.64)\\
		&1.00 &  \textbf{1.00} (0.00)&  0.98 (1.01)&  0.97 (0.00)&  0.98 (1.03)&&  \textbf{12.11} (1.42)&  15.27 (7.57)&  17.58 (3.78)&  16.71 (5.22)&&  \textbf{ 9.59} (1.16)&   8.09 (2.66)&            7.80 (3.59)&   7.64 (1.04)\\
		&1.50 &           1.00 (0.00)&  1.00 (0.00)&  1.00 (0.00)&  1.00 (0.00)&&  \textbf{ 2.91} (9.40)&   4.57 (4.10)&   5.87 (2.17)&   5.16 (2.09)&&  \textbf{31.32} (2.09)&  25.27 (2.85)&           22.82 (0.48)&  23.52 (0.86)\\
		&2.00 &           1.00 (0.00)&  1.00 (0.00)&  1.00 (0.00)&  1.00 (0.00)&&  \textbf{ 0.71} (2.74)&   0.85 (5.56)&   1.34 (6.29)&   1.09 (0.91)&&  \textbf{64.30} (0.67)&  59.21 (1.16)&           54.11 (0.41)&  55.80 (1.25)\\
		\midrule
		\multirow{6}{*}{$\phi=0.9$}
		&0.25 &           0.65 (1.52)&\textbf{0.67} (0.00)&0.54 (1.82)&0.62(1.59)&&         39.38 (0.03)&\textbf{39.13} (4.35)&40.52 ( 3.29)&39.53 (1.15)&&            2.28 (1.30)&            2.84 (0.71)&   \textbf{3.08} (5.12)&   2.68 (1.90)\\
		&0.50 &  \textbf{0.83} (2.47)&         0.82 (3.80)&0.65 (1.56)&0.76(1.33)&&\textbf{32.77} (1.20)&         33.77 (2.46)&36.08 ( 2.78)&35.05 (2.53)&&            3.05 (2.56)&            3.57 (1.92)&   \textbf{3.84} (4.35)&   3.34 (0.00)\\
		&0.75 &  \textbf{0.95} (2.15)&         0.92 (2.22)&0.76 (0.00)&0.85(1.16)&&\textbf{25.39} (4.61)&         26.02 (2.62)&32.63 ( 2.10)&29.23 (2.02)&&            4.69 (2.49)&   \textbf{5.21} (3.37)&            4.78 (4.82)&   4.72 (1.07)\\
		&1.00 &  \textbf{0.99} (1.02)&         0.97 (1.04)&0.84 (0.00)&0.93(1.09)&&\textbf{15.90} (4.26)&         18.95 (0.64)&27.44 ( 1.40)&22.48 (2.60)&&  \textbf{ 8.24} (0.96)&            8.07 (3.07)&            6.48 (1.89)&   7.07 (1.00)\\
		&1.50 &  \textbf{1.00} (0.00)&         1.00 (0.00)&0.94 (0.00)&0.99(1.02)&&\textbf{ 4.71} (7.28)&          6.54 (0.30)&17.29 ( 3.72)&10.52 (0.57)&&  \textbf{26.60} (2.31)&           22.75 (2.99)&           13.79 (1.40)&  18.15 (0.71)\\
		&2.00 &  \textbf{1.00} (0.00)&         1.00 (0.00)&0.98 (0.00)&1.00(1.01)&&\textbf{ 1.38} (2.99)&          1.40 (5.41)& 9.35 (13.33)& 3.65 (1.08)&&  \textbf{57.62} (1.00)&           53.62 (1.38)&           29.31 (2.50)&  42.61 (1.00)\\
		\bottomrule
		\end{tabular}}
	\label{table:resultsCompRef}
\end{sidewaystable}

\newpage
	
	\bibliographystyle{tfcad}
	\bibliography{references}	

\begin{thebibliography}{21}
\newcommand{\enquote}[1]{``#1''}
\providecommand{\natexlab}[1]{#1}
\providecommand{\url}[1]{\normalfont{#1}}
\providecommand{\urlprefix}{}

\bibitem[Adebiyi, Adewumi, and Ayo(2014)]{adebiyi2014comparison}
Adebiyi, Ayodele~Ariyo, Aderemi~Oluyinka Adewumi, and Charles~Korede Ayo. 2014.
  ``Comparison of ARIMA and artificial neural networks models for stock price
  prediction.'' \emph{Journal of Applied Mathematics} 2014.

\bibitem[Arkat, Niaki, and Abbasi(2007)]{arkat2007artificial}
Arkat, Jamal, Seyed Taghi~Akhavan Niaki, and Babak Abbasi. 2007. ``Artificial
  neural networks in applying MCUSUM residuals charts for AR (1) processes.''
  \emph{Applied Mathematics and Computation} 189 (2): 1889--1901.

\bibitem[Chen and Lai(2011)]{chen2011comparison}
Chen, Ling, and Xu~Lai. 2011. ``Comparison between ARIMA and ANN models used in
  short-term wind speed forecasting.'' In \emph{2011 Asia-Pacific Power and
  Energy Engineering Conference}, 1--4. IEEE.

\bibitem[Chen and Yu(2019)]{chen2019deep}
Chen, Shumei, and Jianbo Yu. 2019. ``Deep recurrent neural network-based
  residual control chart for autocorrelated processes.'' \emph{Quality and
  Reliability Engineering International} 35 (8): 2687--2708.

\bibitem[De~la Torre~Guti{\'e}rrez and Pham(2018)]{de2018identification}
De~la Torre~Guti{\'e}rrez, H{\'e}ctor, and Duc~Truong Pham. 2018.
  ``Identification of patterns in control charts for processes with
  statistically correlated noise.'' \emph{International Journal of Production
  Research} 56 (4): 1504--1520.

\bibitem[Elman(1990)]{elman1990finding}
Elman, Jeffrey~L. 1990. ``Finding structure in time.'' \emph{Cognitive science}
  14 (2): 179--211.

\bibitem[Fris{\'e}n(2009)]{frisen2009optimal}
Fris{\'e}n, Marianne. 2009. ``Optimal sequential surveillance for finance,
  public health, and other areas.'' \emph{Sequential Analysis} 28 (3):
  310--337.

\bibitem[Fuqua and Razzaghi(2020)]{fuqua2020cost}
Fuqua, Donovan, and Talayeh Razzaghi. 2020. ``A cost-sensitive convolution
  neural network learning for control chart pattern recognition.'' \emph{Expert
  Systems with Applications} 150: 113275.

\bibitem[Gers, Schmidhuber, and Cummins(2000)]{gers2000learning}
Gers, Felix~A, J{\"u}rgen Schmidhuber, and Fred Cummins. 2000. ``Learning to
  forget: Continual prediction with LSTM.'' \emph{Neural computation} 12 (10):
  2451--2471.

\bibitem[Graves, Mohamed, and Hinton(2013)]{graves2013speech}
Graves, Alex, Abdel-rahman Mohamed, and Geoffrey Hinton. 2013. ``Speech
  recognition with deep recurrent neural networks.'' In \emph{2013 IEEE
  international conference on acoustics, speech and signal processing},
  6645--6649. Ieee.

\bibitem[Heskes(1996)]{heskes1996practical}
Heskes, Tom. 1996. ``Practical confidence and prediction intervals.''
  \emph{Advances in neural information processing systems} 9.

\bibitem[Hochreiter and Schmidhuber(1997)]{hochreiter1997long}
Hochreiter, Sepp, and J{\"u}rgen Schmidhuber. 1997. ``Long short-term memory.''
  \emph{Neural computation} 9 (8): 1735--1780.

\bibitem[Maged and Xie(2022)]{maged2022recognition}
Maged, Ahmed, and Min Xie. 2022. ``Recognition of abnormal patterns in
  industrial processes with variable window size via convolutional neural
  networks and AdaBoost.'' \emph{Journal of Intelligent Manufacturing} 1--23.

\bibitem[Pugh(1989)]{pugh1989synthetic}
Pugh, G~Allen. 1989. ``Synthetic neural networks for process control.''
  \emph{Computers \& industrial engineering} 17 (1-4): 24--26.

\bibitem[Sutskever, Martens, and Hinton(2011)]{sutskever2011generating}
Sutskever, Ilya, James Martens, and Geoffrey~E Hinton. 2011. ``Generating text
  with recurrent neural networks.'' In \emph{ICML}, .

\bibitem[Tran et~al.(2022)]{tran2022application}
Tran, Phuong~Hanh, Adel Ahmadi~Nadi, Thi~Hien Nguyen, Kim~Duc Tran, and
  Kim~Phuc Tran. 2022. ``Application of Machine Learning in Statistical Process
  Control Charts: A Survey and Perspective.'' In \emph{Control Charts and
  Machine Learning for Anomaly Detection in Manufacturing}, 7--42. Springer.

\bibitem[Wang and Zwetsloot(2021)]{wang2021changepoint}
Wang, Zezhong, and Inez~Maria Zwetsloot. 2021. ``A Change-Point Based Control
  Chart for Detecting Sparse Changes in High-Dimensional Heteroscedastic
  Data.''  \urlprefix\url{https://arxiv.org/abs/2101.09424}.

\bibitem[Weese et~al.(2016)]{weese2016statistical}
Weese, Maria, Waldyn Martinez, Fadel~M Megahed, and L~Allison Jones-Farmer.
  2016. ``Statistical learning methods applied to process monitoring: An
  overview and perspective.'' \emph{Journal of Quality Technology} 48 (1):
  4--24.

\bibitem[Wijaya, Kom, and Napitupulu(2010)]{wijaya2010stock}
Wijaya, Yohanes~Budiman, S~Kom, and Togar~Alam Napitupulu. 2010. ``Stock price
  prediction: comparison of Arima and artificial neural network methods-An
  Indonesia Stock's Case.'' In \emph{2010 Second International Conference on
  Advances in Computing, Control, and Telecommunication Technologies},
  176--179. IEEE.

\bibitem[Yu(2019)]{yu2019selective}
Yu, Jianbo. 2019. ``A selective deep stacked denoising autoencoders ensemble
  with negative correlation learning for gearbox fault diagnosis.''
  \emph{Computers in Industry} 108: 62--72.

\bibitem[Zwetsloot et~al.(2023)]{zwetsloot2023remaining}
Zwetsloot, Inez~M., Yu~Lin, Jiaqi Qiu, Lishuai Li, William Ka~Fai Lee, Edmond
  Yin~San Yeung, Colman Yiu~Wah Yeung, and Chris Chun~Long Wong. 2023.
  ``Remaining Useful Life Modelling with an Escalator Health Condition Analytic
  System.''  \urlprefix\url{https://arxiv.org/abs/2306.05436}.

\end{thebibliography}

\end{document}